\documentclass[letterpaper, 10 pt, conference]{ieeeconf} 
\IEEEoverridecommandlockouts
\usepackage{times}
\usepackage[pdftex]{graphicx}
\usepackage{graphics}
\usepackage[cmex10]{amsmath}
\usepackage{url}

\usepackage{array}
\usepackage{tabularx}
\usepackage{multirow}
\usepackage{multicol}
\usepackage{booktabs}
\usepackage{epstopdf}
\usepackage{rotating}
\usepackage{csquotes}

\usepackage[export]{adjustbox}
\usepackage{amssymb}
\usepackage{amsmath}
\usepackage[all]{xy}    
\usepackage{ifthen}
\usepackage[usenames,dvipsnames]{color}
\usepackage{enumerate}
\usepackage{bm}

\makeatletter
\let\NAT@parse\undefined
\makeatother
\usepackage[numbers]{natbib}

\newcommand{\eref}[1]{(\ref{#1})}
\newcommand{\secref}[1]{Section~\ref{#1}}
\newcommand{\tabref}[1]{Table~\ref{#1}}

\newcommand{\figref}[1]{Figure~\ref{#1}}

\newcommand{\myparagraph}[1]{\vspace{0.1in}\noindent\textbf{#1}}

\newboolean{draft-mode}
\setboolean{draft-mode}{true}
\newcommand{\sidenote}[1]{\ifthenelse{\boolean{draft-mode}}{\marginpar{\tiny\raggedright\textsf{\hspace{0pt}#1}}}{}}
\DeclareRobustCommand{\arnote}[1]{\ifthenelse{\boolean{draft-mode}}{\textcolor{blue}{\textbf{AR: #1}}}{}}
\DeclareRobustCommand{\mbnote}[1]{\ifthenelse{\boolean{draft-mode}}{\textcolor{cyan}{\textbf{MB: #1}}}{}}

\title{\LARGE \bf A Probabilistic Data-Driven Model for Planar Pushing}

\author{
    \authorblockN{Maria Bauza and Alberto Rodriguez} 
    \authorblockA{Mechanical Engineering Department --- Massachusetts Institute of Technology\\
        {\tt\small <bauza,albertor>@mit.edu}} 
        \thanks{This work was supported by the National Science Foundation award [NSF-IIS-1637753] through the National Robotics Initiative. Maria Bauza is the recipient of \emph{La Caixa} Fellowship.}}

\begin{document}
\maketitle

\begin{abstract}
  %
  This paper presents a data-driven approach to model planar pushing interaction to predict both the most likely outcome of a push and its expected variability. The learned models rely on a variation of Gaussian processes with input-dependent noise called Variational Heteroscedastic Gaussian processes (VHGP) \citep{Lazaro2011} that capture the mean and variance of a stochastic function.
  We show that we can learn accurate models that outperform analytical models after less than 100 samples and saturate in performance with less than 1000 samples. We validate the results against a collected dataset of repeated trajectories, and use the learned models to study questions such as the nature of the variability in pushing, and the validity of the quasi-static assumption.
\end{abstract}

\section{Introduction}
\label{sec:introduction}

Models of physical interaction in robotics are driven by experimental laws of friction and impact. These laws, such as Coulomb friction, describe the macroscopic behavior of contact by compounding variations at the microscopic level. As a result, one expects them to be accurate at most in a \emph{statistical} sense. We are interested in learning data-driven empirical models that capture more accurately statistical physical interaction, including its expected variability.

\begin{figure}
\includegraphics{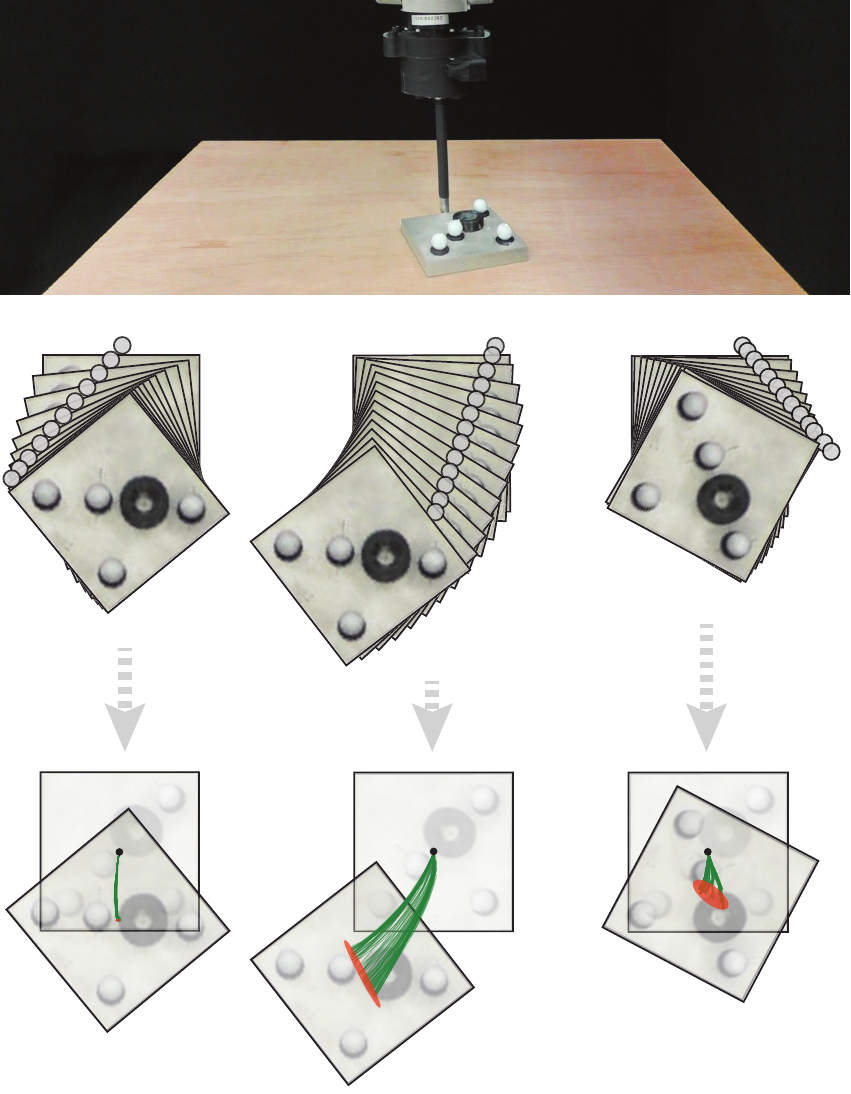}
  \caption{This work learns a data-driven model of the most likely outcome and the expected variability involved in pushing an object. The image shows three different pushes whose outcome (repeated 100 times) yields very different distributions: (left) convergent, (center) divergent, and (right) multi-modal. We show in green the trajectories of the center of mass of the block, and in red an ellipse approximating the distribution of final poses.}
  \label{fig:3_push} 
  \vspace{-0.25in}
\end{figure}

Pushing is a simple manipulation task which already shows interesting statistical behavior. It is primitive to our ability to manipulate objects large and small, and is often involved in more complex manipulation behaviors such as grasping. In previous work~\citep{Yu2016} we provide empirical evidence of the variability in the outcome of a planar push. \figref{fig:3_push} shows that a series of pushes (center), indistinguishable to sensor and actuator resolution, yields divergent outcomes, while a different set of pushes (left) yields a more convergent set of outcomes. Some pushes are more precise than others, and some yield multi-modal behavior (right).

In this paper we learn a compact data-driven model that captures the first two moments, i.e., mean and variance, of the expected behavior of a pushed object. We expect that models like this can be the basis for more realistic simulation, can aid in the design of robust plans or control policies, and can yield more statistically sound inference.
The approach, validation, and structure of this paper, are as follows:
\begin{itemize}
    \item \textbf{Problem:} We are interested in the statistical 
    mechanics of the planar pusher-slider system \citep{Mason1986,Goyal1991,Lynch1996,Hogan2016} where a frictional point contact pushes on a planar object sliding on a frictional surface. 
    \item \textbf{Pushing Data:} We rely on the dataset by \citet{Yu2016} to learn and test the model. We contribute in \secref{sec:pushingData} with an addendum to the dataset with repeated pushes in a grid of pushing locations and directions designed to validate the variances predicted by the model (the new dataset is available online \cite{labwebsite}).
    \item \textbf{Model:} The learned model is based on a family of Gaussian processes called Heteroscedastic Gaussian processes (HGPs), along with their state-of-the-art variational implementation \citep{Lazaro2011}. This model targets phenomena with input-dependent noise, i.e., when the amount of noise introduced by the system depends on the action. \secref{sec:model} uses it to estimate the most likely outcome of a push and its variance.    
    \item \textbf{Evaluation and comparisons:} In \secref{sec:comparisons} we evaluate the push predictions for four objects sliding on four materials. The accuracy is measured by the mean square prediction error (NMSE) and the normalized log probability density (NLPD), and compared to normal Gaussian processes and a common analytical model~\citep{lynch1992manipulation}.
    \item \textbf{Validation:} Finally, in \secref{sec:validation} we validate the predicted probability distributions based on the KL-divergence distance to ground truth estimates of the distribution from repeated pushes. 
\end{itemize}

We finish in \secref{sec:discussion} with a discussion on contributions, limitations, and possible directions for future work.

\section{Related Work}
\label{sec:relatedWork}

There is significant excitement surrounding empirical data-driven techniques for robotic manipulation~\citep{Pinto2016, Agarwal2016}. 
Recently \citet{Huang2016} surveyed efforts to create datasets of object manipulation. The high-fidelity dataset on planar pushing interaction by \citet{Yu2016} is specially relevant to this work. It contains recordings of pushing motions and forces for different dimensions of shape, material, pushing direction, location, velocity, and acceleration. It also provides empirical evidence of the variability of the pushing process, which is the basis of the learned models in this paper.


Over the years, several works have applied data-driven techniques to the problem of planar pushing  \citep{Salganicoff1993,Walker2008,Lau2011,Meric2015,Zhou2016}.
Most recently, \citet{Zhou2016} presented a data-driven but physics-inspired model for planar friction. The algorithm approximates the limit surface representation of the relationship between frictional loads and motion twists at a planar contact, and is the state-of-the-art in data-efficient friction modeling in robotics.
All these algorithms study the problem of controlling a pushed object in a data-driven fashion, but to our knowledge, no previous work has attempted to model both the expected behavior and the experimental variability.

In this paper we use a probabilistic model of the Gaussian processes family to predict the outcome of a push. Gaussian processes are used often to capture both mean and variance of a dynamic system. For example \citet{Paolini2014} use Gaussian processes to learn both the transition dynamics and the observation model of prehensile manipulation tasks. 
In this paper we explicitly consider the dependence of the noise in the transition dynamics with the input action by using heteroscedastic Gaussian processes introduced by \citet{Le2005}. \citet{Kersting2007} proposed a simplified learning algorithm based on maximum likelihood approach, which tends to underestimate noise levels. \citet{Lazaro2011} solve this problem by introducing a variational heteroscedastic Gaussian process algorithm which we use in this work.

\section{Pusher-Slider Data}
\label{sec:pushingData}

\begin{figure}
  \begin{center}
    \includegraphics{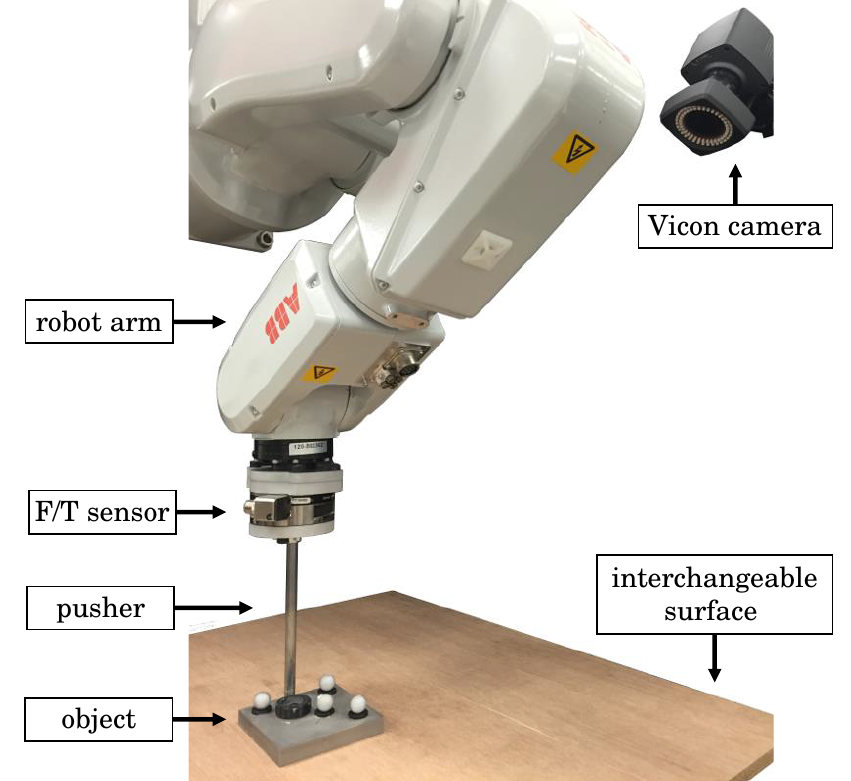}
  \end{center}
  \caption{Experimental setup. The interaction between the pusher (vertical rod) and the object (square block) is recorded through a Vicon tracking system and a force-torque sensor. The system has been designed to provide a clean interface between the pusher and slider, and is described in detail in \citet{Yu2016}.}
  \label{fig:hardware_robot}
\end{figure}

\begin{figure}
  \includegraphics{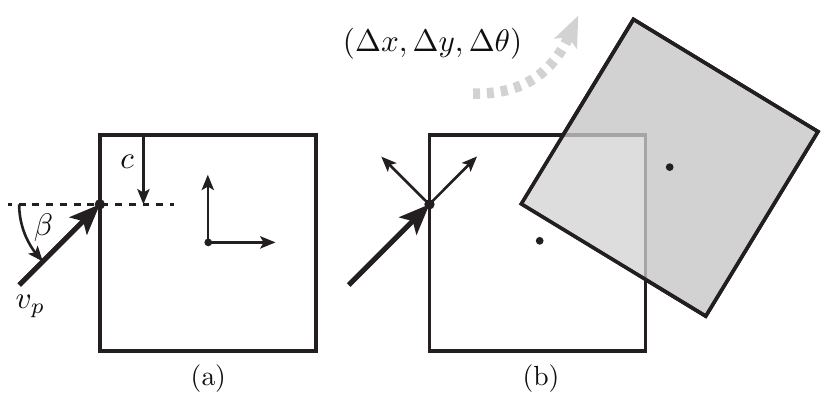}
  \caption{Input/output space for the dynamics of the pusher-slider system. (a) The input is parametrized with the distance $c$ that captures the contact point along an edge of the square (for objects without symmetries, contact points are evenly spaced along their perimeter), the angle of push $\beta$, and its velocity $v_p$. (b) The output is parametrized with $(\Delta x, \Delta y, \Delta \theta)$, the displacement of the object in a reference frame aligned with the push direction.}
  \label{fig:inputs&outputs}
  \vspace{-0.15in}
\end{figure}

This paper focuses on modeling the probabilistic behavior of the pusher-slider system, illustrated in \figref{fig:hardware_robot}. This section describes the data we use to learn and validate the model.
We are interested in learning the behavior of the slider as an input-output relationship. As illustrated in \figref{fig:inputs&outputs} the representation for the input space is:
\begin{itemize}
    \item[]$\boldsymbol{v_p}$ Magnitude of the velocity of the pusher.
    \item[]$\boldsymbol{c} $ Contact point on the perimeter of the slider.
    \item[]$\boldsymbol{\beta} $ Pushing angle.
\end{itemize}
and the representation for the output space is:
\begin{itemize}
    \item[] $\boldsymbol{\Delta x}$ COM x displacement in the pusher ref. frame.
    \item[] $\boldsymbol{\Delta y}$ COM y displacement in the pusher ref. frame.
    \item[] $\boldsymbol{\Delta \theta}$ Orientation change.
\end{itemize}
after a push for $\Delta t$ seconds. These parameters are sufficient to characterize simple models of planar point pushing~\citep{lynch1992manipulation,Hogan2016}. 


In this paper we use two sets of data: a large-scale general purpose dataset of planar pushing for learning, and a dataset with repeated pushes we collected for the purpose of validating the results. In particular we validate the model prediction of the variance for the outcome of a push.

\subsection{Learning Data}
\label{learning_data}
In previous work~\citep{Yu2016} we captured a large set of planar pushes with the robotic setup in \figref{fig:hardware_robot}, composed of a high-precision industrial robot fitted with a cylindrical pusher and stainless steel objects of different shapes sliding on surfaces of different materials.
The system records the trajectories of the pusher and object and the force at the interaction. 
%
We will show in later sections that 100 samples are sufficient in average to outperform analytical models, and model accuracy saturates with less than 1000 samples.

\subsection{Validation Data}
\label{subsec:validation_data}
Our goal is to predict reliably the object's expected motion and its variance:
\begin{align}
\Delta x  &\sim \mu_x(u), \sigma_x^2(u) \nonumber\\
\Delta y  &\sim \mu_y(u), \sigma_y^2(u) \\
\Delta \theta &\sim \mu_{\theta}(u), \sigma_{\theta}^2(u) \nonumber 
\end{align}
%
%
where $u = (v_p,c, \beta)$ is the input/action and $\mu(u)$ and $\sigma^2(u)$ are the \emph{input-dependent} expected outcome and variance. The dependence of the variance $\sigma^2(u)$ with the input 
$u$ is the key complexity we address in this paper, motivated by the example in \figref{fig:3_push}, and leads us to consider HGPs instead of standard GPs.

To validate the observation that the output noise depends on the input, we collected a new dataset in the same setup in \figref{fig:hardware_robot} containing 100 repetitions of each push considered, which gives us an approximate distribution of the object motion.
%
%
In this new dataset, the pusher follows a straight trajectory of 1cm long at 20mm/s. The initial contact angles go from $-1.5$ to $+1.5$ radians spaced by 0.1 radians while we consider 11 different initial contact points evenly spaced on the side of the object. This produces a sufficiently dense grid of pushes allowing us to extract for each push the expected mean and variance of the object motion. This dataset is available online \cite{labwebsite}.

Besides validating the proposed model, the new dataset also provides empirical evidence to study the variability of the pushing process, including multi-modality effects.

\section{Heteroscedastic Gaussian processes}
\label{sec:HGP}

Gaussian processes (GPs) are a flexible and commonly used framework in robotics. We find them in many different areas in robotics such as manipulation, motion planning or learning from demonstration \citep{Paolini2014,Mukadam2016,Choi2016}. In this paper, we apply GPs to the modeling of planar pushing. 

Classical (homoscedastic) GPs~\citep{Rasmussen2006} assume that the noise from the observations is Gaussian and constant over the input:
\begin{equation}
    y(x) = f(x) + \varepsilon
\end{equation}
where $y(x)$ is an observation of the process at the input $x$, $f(x)$ is the latent function that we want to regress and $\varepsilon \sim N(0, \sigma^2)$ represents Gaussian noise with variance $\sigma^2$.

The assumption of constant Gaussian noise together with a GP prior on the latent function $f(x)$ makes analytical inference possible for GPs \eref{eq:GPdistribution}. We consider that $f(x)$ follows a GP prior $f(x) \sim GP(0, k(x,x'))$ where $k$ is the Automatic Relevance Determination Squared Exponential (ARD-SE) kernel \citep{Rasmussen2006}.
Given the training set $D = {\{(x_i, y_i)\}_{i=1}^n}$, the probability of an observation $y_*$ at the input $x_*$ is given by:
\begin{align}\label{eq:GPdistribution} 
  p(y_* | x_*, D, \alpha) &= N(y_*| a_*, c^2_* + \sigma^2)   \nonumber \\
  a_* &= k_*^T(K+\sigma^2I)^{-1}y \\
  c^2_* &= k_{**} - k_*^T(K+\sigma^2I)^{-1}k_* \nonumber
\end{align}
where $K$ is a matrix that evaluates the ARD-SE kernel in the training points, $[K]_{ij} = k(x_i, x_j)$, $k_*$ is a vector with $[k_*]_i = k(x_i, x_*)$ and $k_{**}$ is the value of the kernel at $x_*$, $k_{**}= k(x_*, x_*)$. Finally, $y$ represents the vector of observations from the training, and $\alpha$ is the set of hyperparameters including $\sigma^2$ and the kernel parameters that are optimized during the training process. 

The expected variance of an observation $y_*$ at the input $x_*$ comes from the addition of two variances: $\sigma^2$ and $c_*^2$ \eref{eq:GPdistribution}. While $\sigma^2$ is constant and represents the process noise, $c_*^2$ depends on $x_*$ and is only related to the regression error (not to the real process from which the data originates). In practice, the term $c_*$ depends highly on the density of training points around the input $x_*$ and reflects how confident is the GPs regression on the mean value $a_*$.

Assuming that the process noise $\sigma^2$ is constant over the input space is sometimes too restricting. Allowing some input regions to be more noisy than others is beneficial for practical applications where there are naturally stable and unstable dynamics. As we motivated in the introduction, this is the case for pushing where depending on the type of push, the resulting motion can be convergent or divergent (\figref{fig:3_push}).

Some extensions of GPs can incorporate input-dependent noise. These algorithms are referred as heteroscedastic Gaussian processes (HGPs) and can regress both the mean of the process and its variance over the input space. Moreover, they also improve the mean estimation by giving more relevance to those observations obtained from less noisy inputs.
In HGP regression, observations are drawn from:
\begin{equation}
    y(x) = f(x) + \varepsilon(x)
\end{equation}
where $\varepsilon(x) \sim N(0, \sigma^2(x))$ now explicitly depends on $x$. 

In this paper, we use the state of art algorithm for HGPs called variational HGP (VHGP) proposed by Lazaro-Gredilla and Titsias \citep{Lazaro2011}. Using Bayesian variational theory they derive closed-form solutions for the mean and variance of the process considering input dependent noise. As a result, the probability of an observation $y_*$ is given by:
\begin{align}
    p(y_* | x_*, D, \alpha) &= N(y_*| a_*, c_*^2 + e^{b_*+d_*^2/2}) \nonumber \\     
    a_* &= k_{f*}^T(K_f+R)^{-1}y   \nonumber \\
    c_*^2 &= k_{f**} - k_{f*}^T(K_f+R)^{-1}k_{f*}  \\
    b_* &= k_{g*}^T(\Lambda - \frac{1}{2}I)1 + \mu_0  \nonumber \\
    d_*^2 &= k_{g**} - k_{g*}^T(K_g - \Lambda^{-1})^{-1}k_{g*} \nonumber
\end{align}
where $k_f$ and $k_g$ are the ARD-SE kernels of the mean $f(x)$ and the logarithm of the variance $g(x) = \log \sigma^2(x)$. The matrix $R$ is diagonal with $[R]_{ii} = \sigma^2(x_i)$, $\mu_o$ is the hyperparameter of the log-variance mean and $\Lambda$ is a positive semidefinite diagonal matrix optimized together with the other hyperparameters using conjugate gradient descent.

The similarity between the inference equations for GPs and VHGPs is remarkable. The term $c_*^2$ makes equally reference to the mean error due to the regression process and the equation for $a_*$ is essentially the same. The constant noise $\sigma^2$ in GPs however is substituted by the input-dependent expression $e^{b_*+d_*^2/2}$, the expected noise of the process at the input $x_*$. In terms of computational cost, GPs and VHGPs scale alike with the amount of training data.












\section{The learned model}
\label{sec:model}

\begin{figure}
\includegraphics[width=3.2in,trim={0 0in 0 0in}]{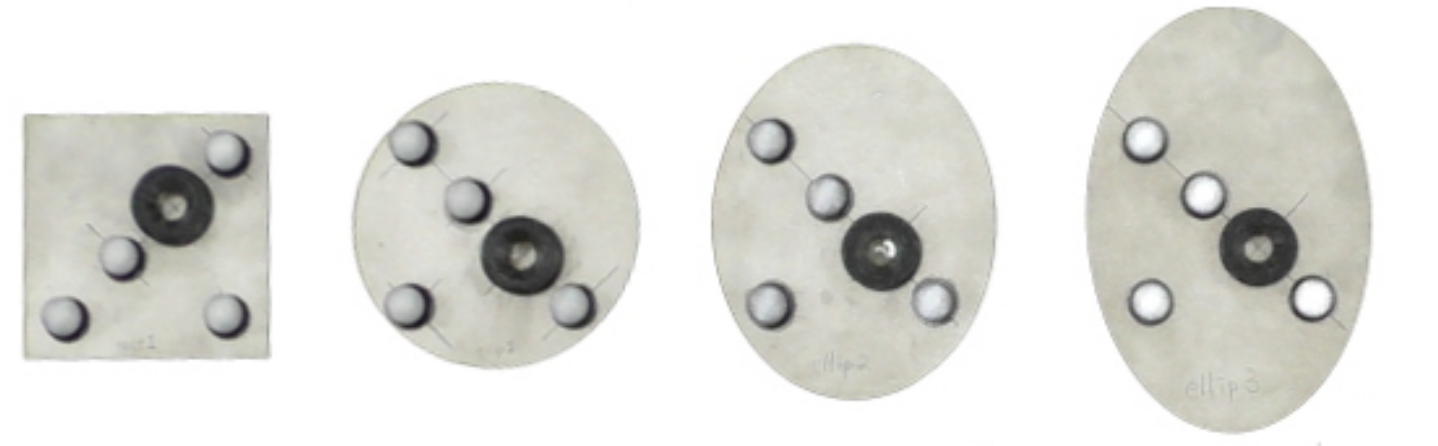}
\caption{Objects used for learning, from \citet{Yu2016}.}
\vspace{-0.10in}
\label{fig:objects}
\end{figure}

\begin{figure*}
  \begin{center}
    \includegraphics[width=6.9in,trim={0 0.5in 0 0.2in}]{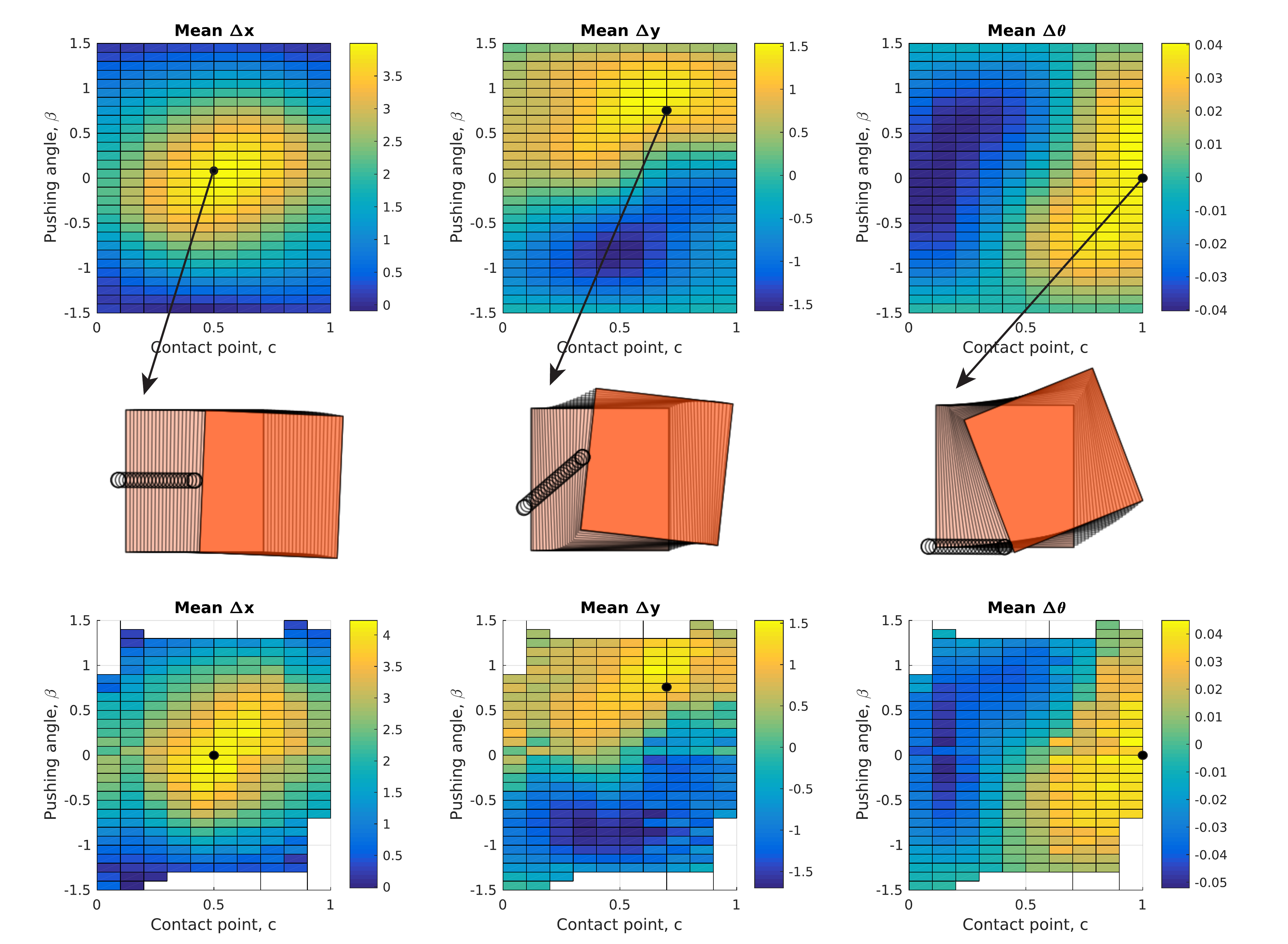}
  \end{center}
  \caption{\textbf{Expected displacement after a push (units: mm and rad).} The first row shows the predicted outputs by the learned VHGP model for a fixed velocity $v_p=20$mm/s after a push of $\Delta t = 0.2$ sec, for different contact point locations and push angles. The result is remarkably similar to the expected outputs from the validation data in the 3rd row. The symmetries in the plots are due to the symmetries of the square. 
  The second row shows pushes that are predicted to yield high displacement in $\Delta x$ ($c = 0.5$, $\beta = 0$), $\Delta y$ ($c=0.7$, $\beta = 0.7$), and  $\Delta \theta$ ($c=1.0$,  $\beta=0$) correspondingly. 
  %
  %
  } 
  \label{fig:colormap_outputs_mean}
\end{figure*}

\begin{figure*}
  \begin{center}
    \includegraphics[width=6.9in,trim={0 0.5in 0 0.2in}]{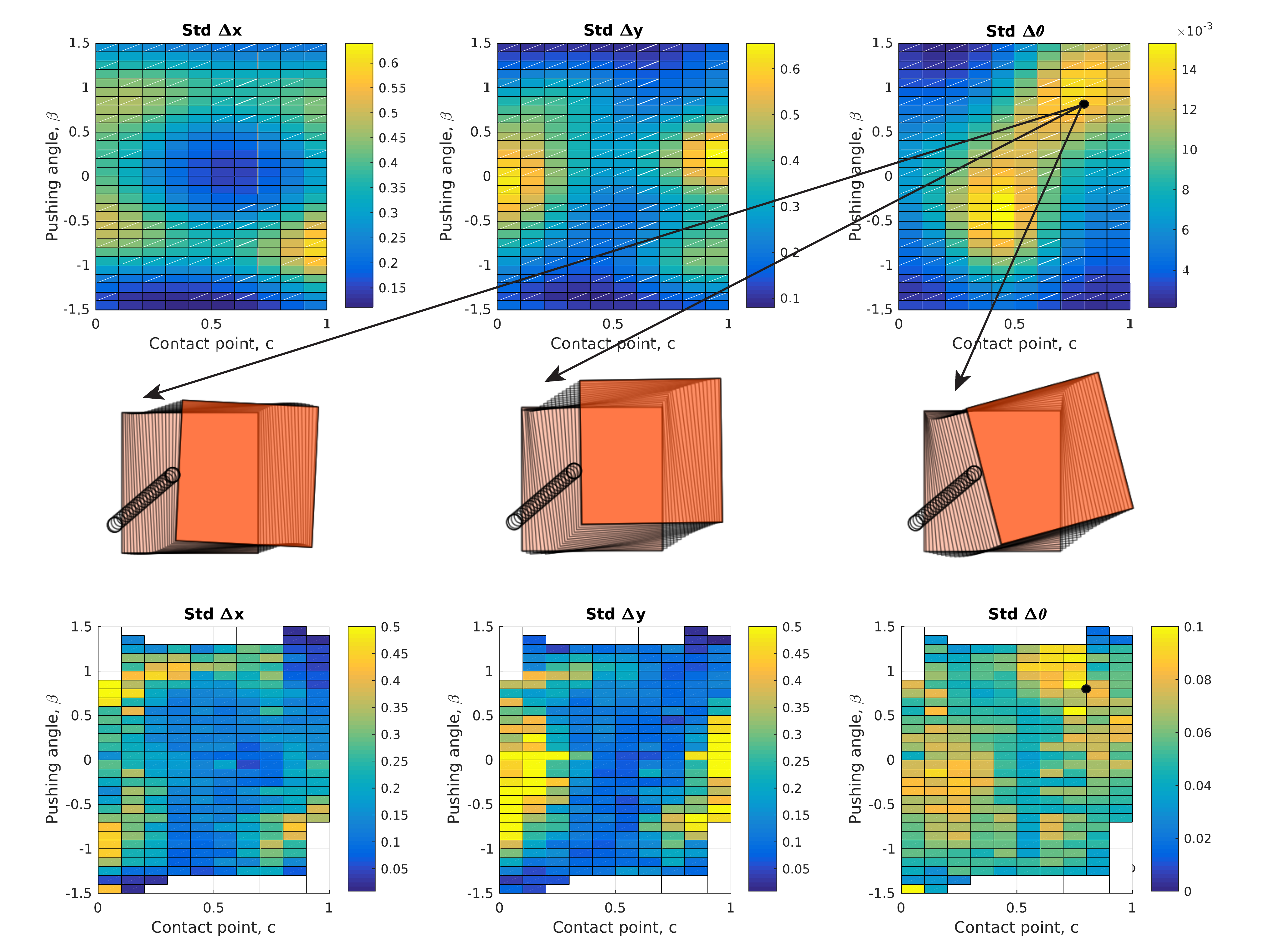}
  \end{center}
  \caption{\textbf{Expected variability after a push (units: mm and rad).} The first row shows the predicted standard deviation (std) of each output obtained from the VHGP model, as compared to the observed from the validation dataset in the third row.
  Some regions are of high variance and from our experience these seem to be independent of the type of surface considered, suggesting that some pushes are more stable than others regardless of the surface. The middle row shows three examples from the learning data where the push considered has the maximum variance in the change of orientation. The outcome does seem to vary considerably.}
  \label{fig:colormap_outputs_variance}
\end{figure*}

In this section we describe in more detail the process to learn the pushing model for the four objects in \figref{fig:objects} sliding on a horizontal surface of four different materials.

To compute the pushing model of a certain pair object-surface, we train three independent VHGPs, one for each output $\Delta x$, $\Delta y$, and $\Delta \theta$. While not optimal from a data-efficiency perspective, since we neglect the existing correlations between the outputs variables, it proves sufficient in terms of performance as long as enough training data is used. For future work it would be interesting to combine multi-task GP prediction with heteroscedastic GPs.

Once each VHGP is trained, we visualize the effect of different pushes given a fixed velocity. Each push is defined by the contact point $c$ and the pushing angle $\beta$. \figref{fig:colormap_outputs_mean} shows the regressed distribution for the case where the pusher's velocity is $v_p=20$mm/s, the object shape is a square, and the surface is plywood. In accordance with intuition, we see that the displacement in the pusher's direction $\Delta x$ is maximum when pushes are done at the center of the edge, $c = 0.5$, and perpendicular to the edge, $\beta = 0$. The maximum change in orientation happens when pushing in between the edge center and the vertex with a pushing angle of $\beta = +30^o$ if $c = 0.75$ and $\beta = -30^o$ if $c = 0.25$.

Analogously, \figref{fig:colormap_outputs_variance} shows the modeled and experimental standard deviation of the predicted pushes, as a function of the contact location and direction. We observe that the magnitude of the noise is in between $10\%$ and $40\%$ of the magnitude of the expected output, which is significant, and further motivates the work in this paper.

There are well defined regions that present more noise than others. To a certain degree the prediction matches well with the validation data, also shown in \figref{fig:colormap_outputs_variance}. 
The precision of the measurement equipment (Vicon tracking system), suggests that the regressed noise comes principally from the pushing process and not from sensor noise. It is further indicative that the shape of the regressed noise remains more o less constant when considering the same object but a different surface. 


A data-driven model that captures the uncertainty of interaction, beyond the deterministic predictions of standard analytical models gives us a more complete perspective of the dynamics of pushing.
This information can be used to differentiate between more and less stable pushes, and improve multi-step prediction, by propagating uncertainty.

\section{Evaluation of the Learned Model}
\label{sec:comparisons}
\label{sec:evaluation}
We evaluate the performance of the learned model with the standard metrics normalized mean square error (NMSE) and normalized log probability density (NLPD):
\begin{align} \label{eq:NMSE}
    \textnormal{NMSE} &= \frac{\sum_{j=1}^{m}(y_j-\hat{y}_j)^2 }{\sum_{j=1}^{m}(y_j-\bar{y})^2}\\
    \textnormal{NLPD} &= -\frac{1}{m}\sum_{j=1}^m \log p(y_j | D) \label{eq:NLPD}
\end{align} 
%
where $m$ is the number of elements in the test set and $\{{y}_j\}_{j=1}^m$ the observations. We note with $\hat{y}_j$ the predicted value from the VHGP model at $x_j$ and with $\bar{y}$ the mean of the observations in the training set. We consider the total NMSE of the model as the sum of the individual NMSEs for each output. For the total NLPD, we consider  $p(y_j|D)$ as the product of the individual probabilities of each output given their respective VHGP.

Note that NMSE, as defined in \eref{eq:NMSE}, does not take into account the variability of motion and only considers its expected value . It reflects the squared distance between the model outputs and the real process normalized by the variance of the observations. The NMSE is especially useful when comparing the results with deterministic models of pushing. The NLPD metric instead, computes how likely are the observations according to a given probabilistic model. It is more appropriated for evaluating our data-driven models as they also regress the uncertainty of the motion in \eref{eq:NLPD}. For the NLPD, lower results imply a higher likelihood of the data. Consequently NLPD penalizes both underconfident and overconfident models, so deterministic models can not be fairly evaluated using NLPD.

\begin{figure}
  \begin{center}
    \includegraphics[width=6.4in,trim={6in 3in 0 2.2in}]{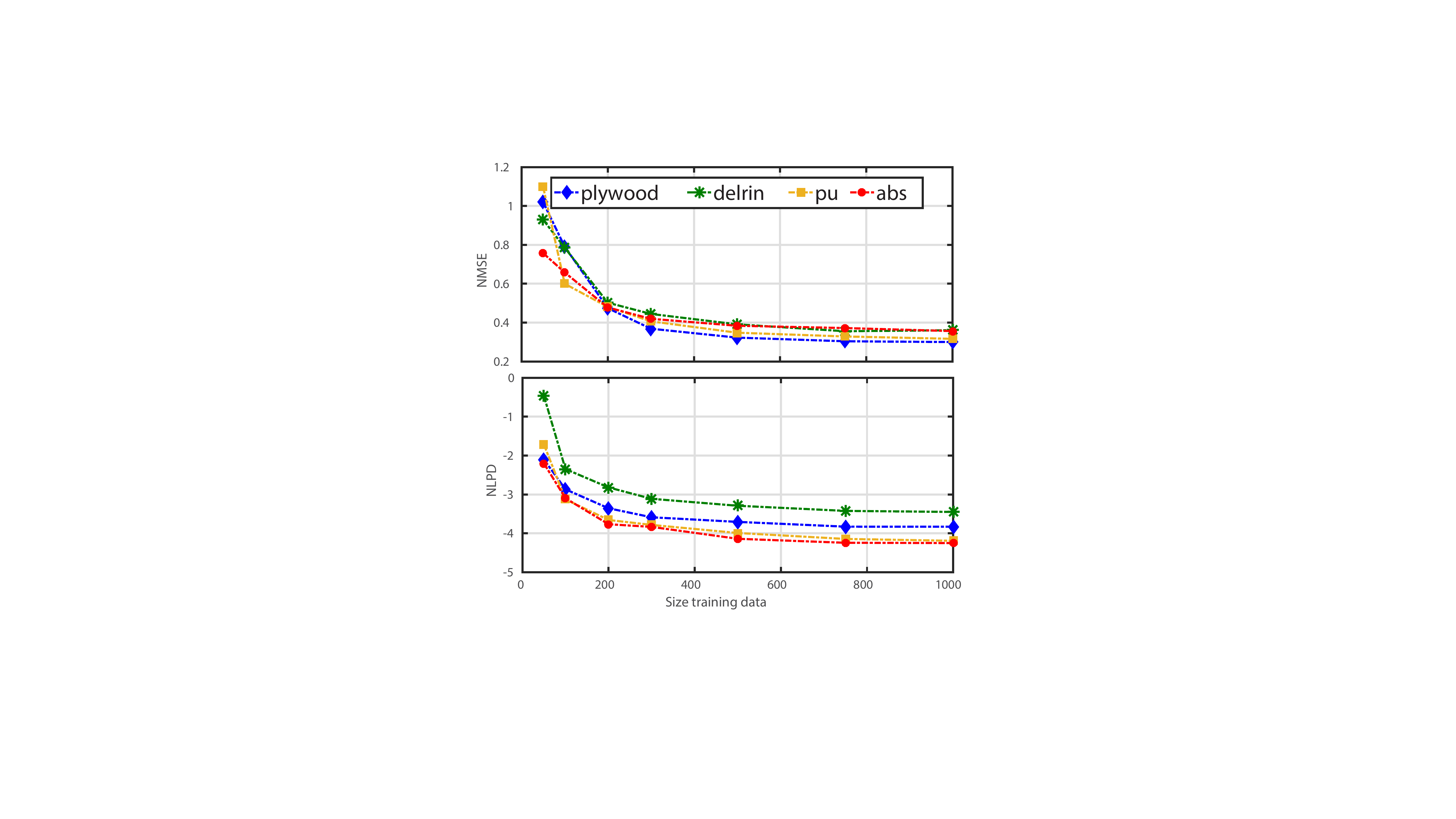}
  \end{center}
  \caption{VHGP model evaluated on the four surfaces used in \citep{Yu2016} depending on the size of the training set. We observe that the model is not very sensitive to the type of surface considered as the errors in NMSE and NLPD are reasonably similar. Higher NLPD in $delrin$ and $plywood$ could be due to higher degradation with time.}
  \label{fig:data_vs_surface}
\end{figure} 

\begin{figure}
  \begin{center}
    \includegraphics[width=6.4in,trim={6in 3in 0 2.2in}]{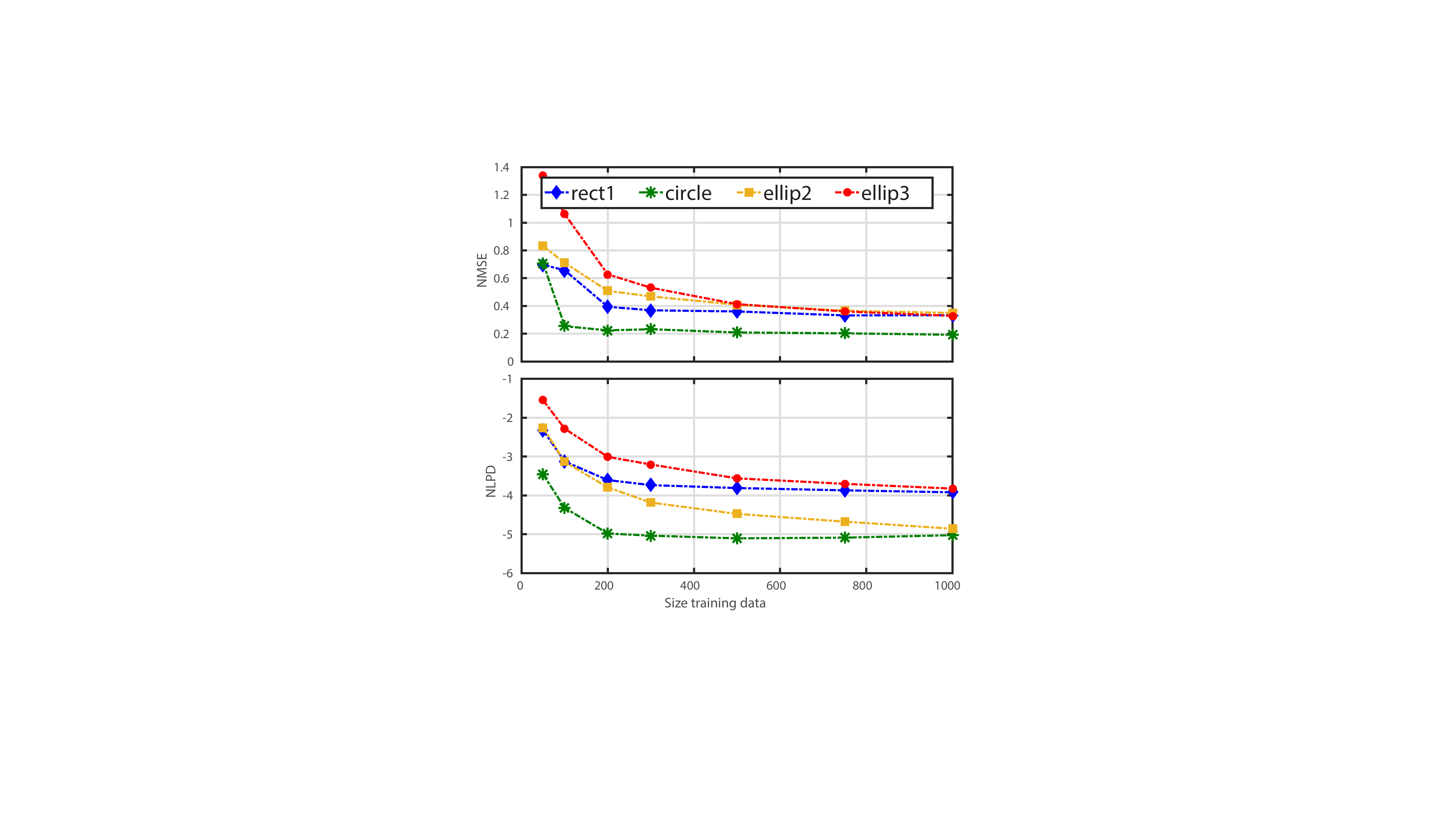}
  \end{center}
  \caption{Evaluation of the VHGP model for the  objects in \figref{fig:objects}. We observe that the circle is easier to learn than the others shapes. This is probably due to the fact the all contacts points in a circle are indeed equivalent. Thus the contact point dimension can be considered redundant, simplifying the training of the model.}
  \label{fig:data_vs_shape}
\end{figure} 

\subsection{Evaluation on Different Objects and Materials}
\figref{fig:data_vs_surface} and \figref{fig:data_vs_shape} show the evaluation of the model when trained on four different surfaces (plywood, Delrin, polyurethane, and ABS) and for four different object shapes (square, circle and two ellipses, as shown in \figref{fig:objects}).
An advantage of VHGPs models is that they do not require a large amount of data. The plots show that less than 1000 samples are sufficient to saturate performance, which is equivalent to collecting about 5 minutes of pushing experiments in our setup if we use $\Delta t = 0.2$s. Therefore our VHGP models can provide good planar push models of new objects and surfaces without having to go through an extensive data collection. 
The fact that VHGPs performance remains more or less constant after a certain amount of data also suggests that the pushing motion is a sufficiently well defined process and simple enough to be learned in general from a reduced exploration of the environment.

\subsection{Comparison with Other Models}

\begin{figure}
  \begin{center}
    \includegraphics[width=6.4in,trim={6in 3in 0 2.2in}]{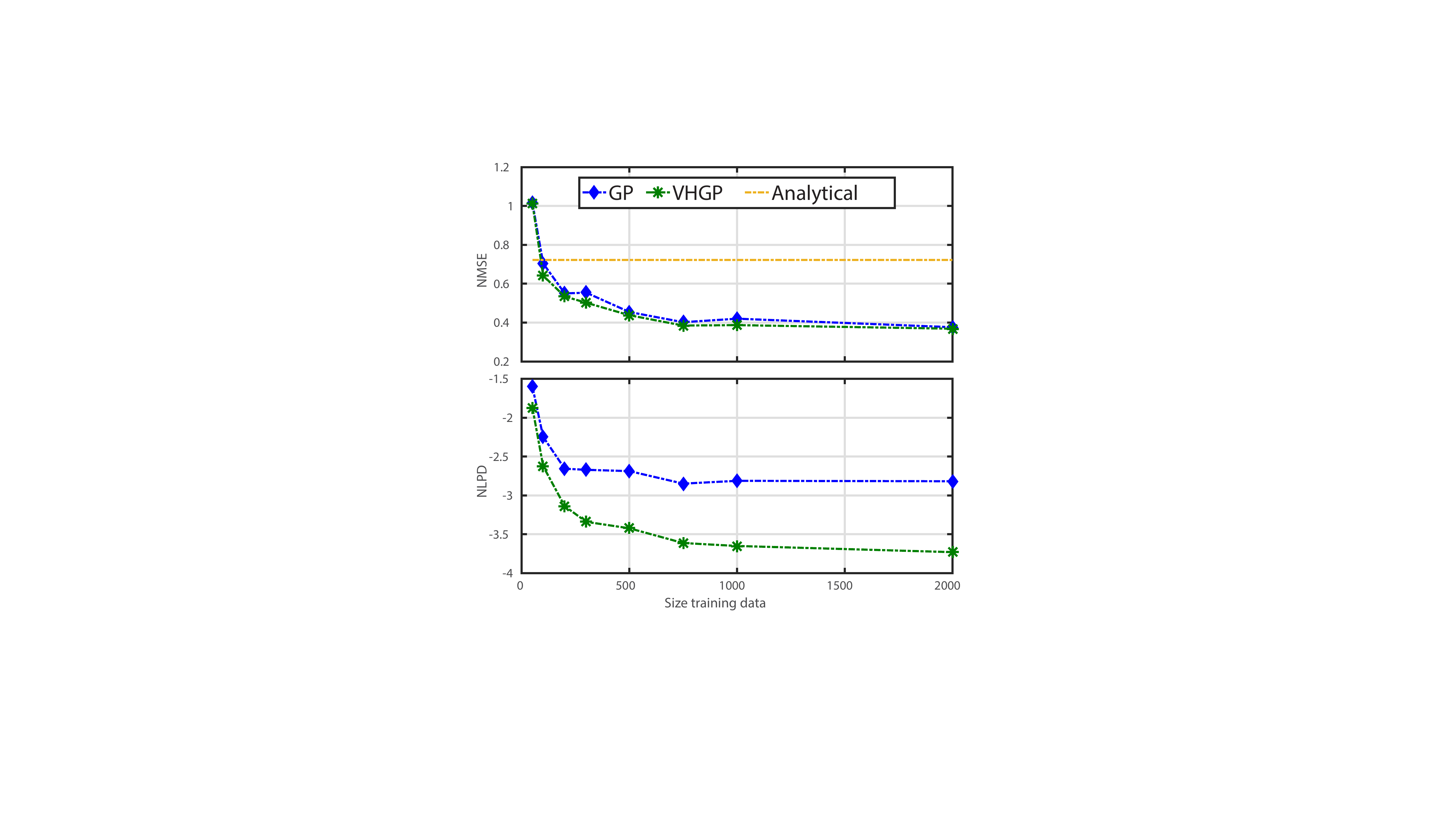}
  \end{center}
  \caption{NMSE and NLPD comparison for different models as a function of the training data. Both GPs and VHGPs outperform the analytical model after 100 samples. As expected, VHGP outperforms GP measured with NLPD, since it has more flexibility to capture the stochasticity of the data.}
  \label{fig:NLPD_models_comparison} \label{fig:NMSE_models_comparison}
\end{figure}

\begin{table}
  \caption{NMSE and NLPD comparisons between models}
  \label{tab:NMSE} \label{tab:NLPD}
  \centering
    \begin{tabular}{| l | l | l | l |}
    \hline
    Outputs & Analytical & GP  & VHGP \\ \hline
    NMSE &  0.72 & 0.38   & 0.37 \\ \hline
    NLPD & \quad -- & -2.82 &   -3.73 \\ 
    \hline
    \end{tabular}
\end{table}

In this section we compare the performance of VHGP against a standard GP model, and a commonly used analytical model~\citep{lynch1992manipulation}.
%
GP and VHGP show a similar NMSE. This is expected since NMSE only evaluates the most likely outcome, and both GPs and VHGPs have a very similar formulation for the regressed mean. The difference in NLPD however, justifies the need of input-dependent noise to explain the stochastic behaviour of pushing. As all inputs in a classical GP are supposed to have the same noise, all the observations become equally weighted during training. Instead, in VHGPs those observations from lower noise regions have higher relevance in the regression process. 

We also compare our data-driven model with the analytical model proposed by \citet{lynch1992manipulation}, that relies on assumptions of quasi-static, uniform pressure distribution, uniform coulomb friction, and an ellipsoidal approximation to the limit surface.



The proposed data-driven model, instead of relying on a perfect knowledge of the object-surface interaction, outputs a distribution of the motion's outcome that encloses unexpected behaviours. 
\figref{fig:NLPD_models_comparison} shows that both GPs and VHGPs outperform the analytical model after 100 samples approximately. 
%

We also observed that for high velocities the analytical model is more unreliable. This is reasonable as the analytical model assumes a quasi-static interaction and does not take into account inertia. In our model, those dynamic effects are captured by adding the pusher velocity as an input and through the uncertainty of the distribution. 

\section{Validation}
\label{sec:validation}

The training and testing set used from~\citep{Yu2016} does not contain sufficient repeated pushes to estimate ground truth variance for a given push.
To validate the distributions predicted by our model, we captured a benchmark dataset that incorporates repeated trajectories so that a reliable notion of uncertainty can be extracted from them (the dataset is available in \cite{labwebsite}). The differences between repeated pushes should mainly include the uncertainty in the object-surface interaction, i.e., the stochastic side of the push itself. Given the samples of each repeated push, we compute the expected motion of the object and its variance from the benchmark. \secref{subsec:validation_data} contains a more detailed description of the dataset, and the bottom rows of \figref{fig:colormap_outputs_mean} and \figref{fig:colormap_outputs_variance} illustrate the obtained means and variances.

To evaluate the results of the model learned from the training dataset in \citep{Yu2016}, against the benchmark dataset, we use the average KL divergence over the pushes. If we assume that the real distribution of each push comes from three independent Gaussians, we can directly use the KL divergence between two Gaussian distributions:
\begin{equation}
    KL(p,q) = \frac{1}{2} \left(log\left(\frac{\sigma_2}{\sigma_1}\right) + \frac{\sigma^2_1 + (\mu_1-\mu_2)^2}{\sigma_2^2} -1 \right)
\end{equation}
where $p= N(\mu_1, \sigma_1^2)$ and $q= N(\mu_2, \sigma_2^2)$.

The average and median KL divergences of each model show that our model not only regresses properly the expected motion, but also can predict reasonably well its distribution (\tabref{tab:KL_divergence}). This is crucial to create robust probabilistic models that can be latter incorporated into multi-step planning and control.

\begin{table}
  \caption{KL divergence between the validation data and the models}
  \label{tab:KL_divergence}
  \centering
  \begin{tabular}{|l|l|l|l|} \hline
    \bfseries  KL divernge & \bfseries GP & \bfseries VHGP \\ \hline
     Average KL   & 22.63 & 15.32 \\ \hline
     Median KL  & 5.74 & 5.34  \\ \hline
  \end{tabular}
\end{table}

\section{Discussion}
\label{sec:discussion}

In recent work~\citep{Yu2016,Kolbert2016} we provided empirical evidence that planar frictional interaction shows non-trivial statistical behavior, and suggested that a probabilistic model might yield a practical and less over-confident approach to model frictional contact. 
This paper is a step in that direction. We focus on the problem of planar pushing, which has proven essential for many types of interaction, both simple and complex, and for which the robotics community has developed a good and long-standing analytical understanding.

\myparagraph{Input-dependent noise.} This work starts from the empirical observation that the magnitude of the observed variability under a constant push can vary up to an order of magnitude with the pushing action, i.e., pushing location, pushing direction or pushing velocity. 
A model that accounts for that variability could be used, for example, to avoid actions that yield unpredictable behavior.

\myparagraph{Data-driven modeling.} Building from the recent pushing dataset by \citet{Yu2016}, we propose to use a data-driven modeling approach based on a Variational Heteroscedastic Gaussian process model (VHGP) to capture the mean and variance of a planar frictional push. 
Unlike traditional Gaussian processes, which assume a level of output noise independent of their input, heteroscedastic Gaussian processes model the dependence of both the mean and variance of the outcome of a planar push as functions of the input pushing action. 

The learned models are specific to the particular object and material. Generalization over materials and shapes is an interesting question that we plan to explore in future work, which will require a model less computationally expensive to train, such as those based on sparse Gaussian processes \citep{Snelson2006} or  Random Features \citep{Rahimi2008}.

\myparagraph{Evaluation.} We show that an order of 100 samples is sufficient to overcome the performance of a standard analytical model, and evaluate the improvement of the VHGP framework over a traditional homoscedastic GP. The accuracy of the VHGP model tails off at about $10^3$ training samples, data that can be captured in about 5 minutes in our training setup.

We validate the model against a new dataset collected in the same setup as~\citep{Yu2016} for the purpose of benchmarking. This new dataset is composed of 100 identical pushes for each of a set of more than 300 different pushing actions, which gives an empirical footprint of the stochasticity of planar pushing. The performance, evaluated by means of the KL divergence shows a clear improvement over a normal GP. 

\begin{figure}
  \begin{center}
    \includegraphics[width=3.8in,trim={6.2in 3.2in 5in 2.8in}]{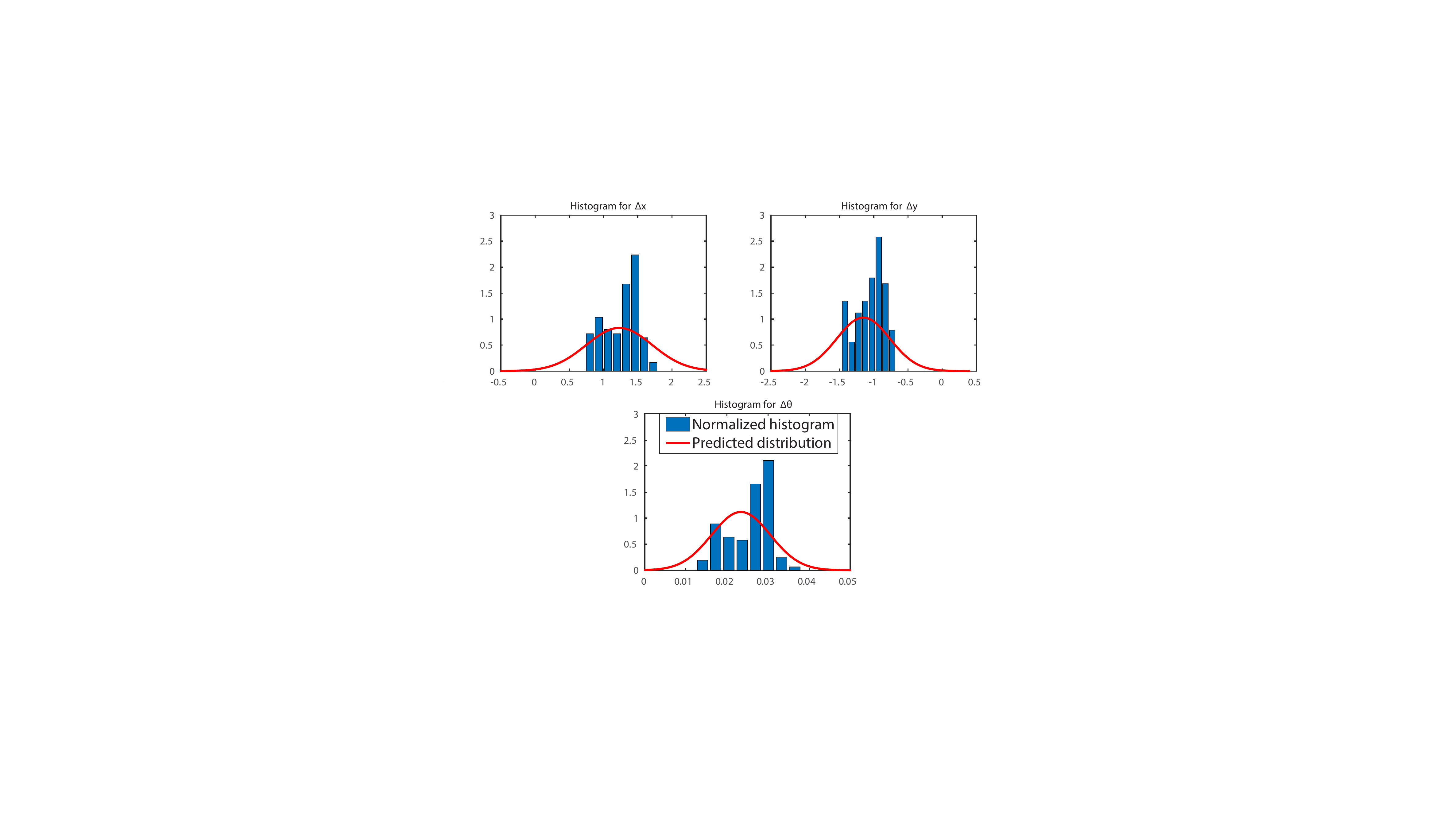}
  \end{center}
  \caption{These histograms represent the different outputs obtained when repeating the same push. They have been normalized so that they can be compared with the distribution provided by our VHGPs. In this case, distributions do not adjust clearly to a Gaussian distribution but our model still obtains a good approximation for them.}
  \label{fig:histograms_for_modality}
\end{figure} 

\myparagraph{Gaussianity vs. multi-modality.} While VHGPs are the state-of-the-art for data efficient regression for input dependent processes, Gaussianity and unimodality of the underlying dynamics is still a key assumption. We know that this is not always the case (\figref{fig:histograms_for_modality}). 
However, a model with input-dependent noise can express multi-modal behavior when integrated over time. We are interested in exploring further the idea of capturing finite multi-modal behavior with uni-modal Gaussian instantaneous models. 

\myparagraph{Dependence with velocity.} A very common assumption in robotic manipulation is to neglect the inertial effects of interaction, i.e. the quasi-static assumption.
In this paper we avoid that assumption by making the velocity of the pusher an explicit input to the model.

\begin{figure}[t]
  \begin{center}
    \includegraphics[width=4in,trim={6.2in 4in 5in 4in}]{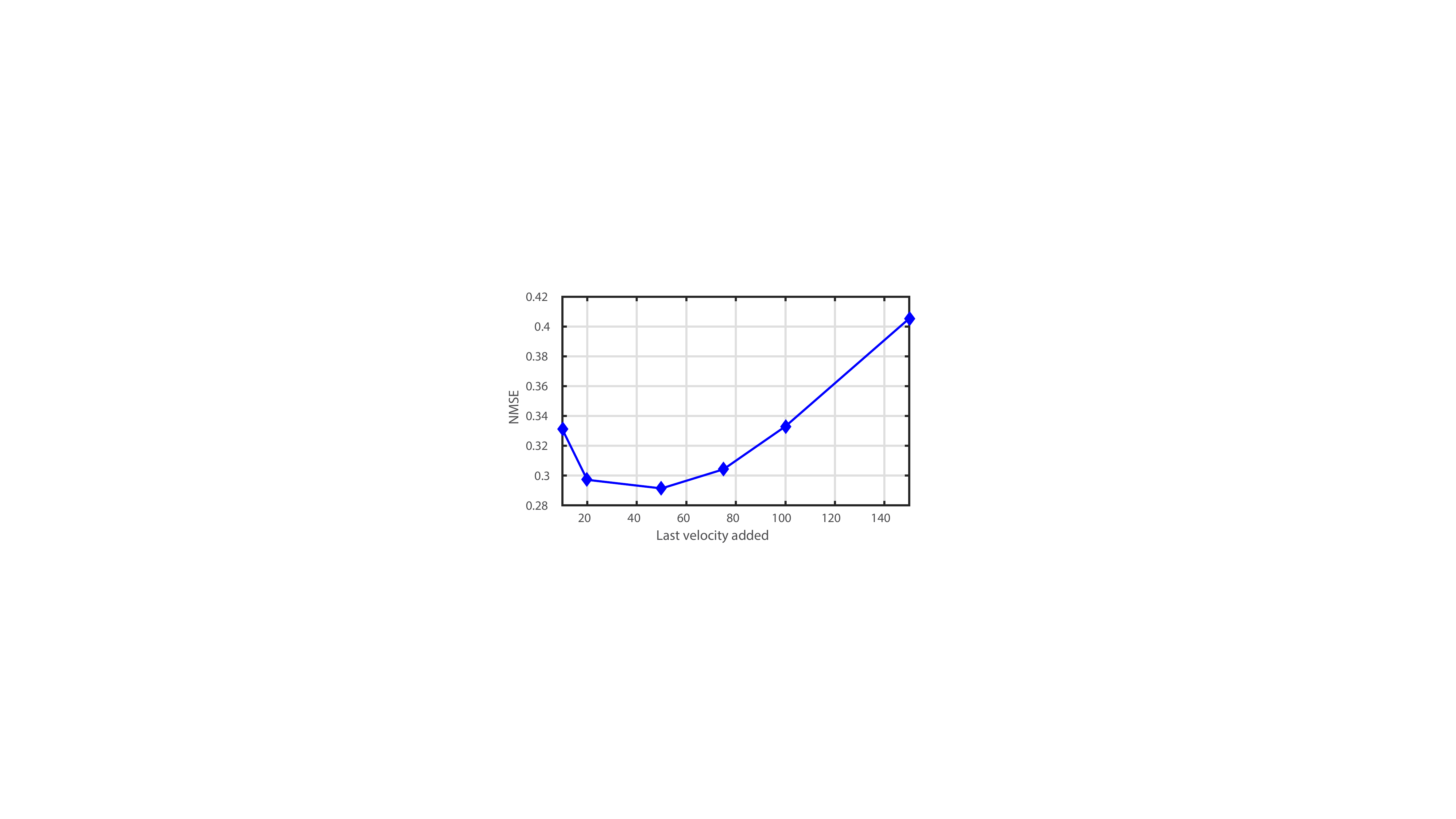}
  \end{center}
  \caption{The x-axis of the plot shows the max velocity added to train a pushing model without velocity as an input. The plot shows the evolution of the performance (NMSE) of these velocity-independent models trained for a larger bracket of velocities in the training data. Initially, adding more velocities improves performance. However, after 50-80mm/s the performance of the model worsens. We conclude that the quasi-static assumption does not hold anymore after that.}
  \label{fig:quasi-static}
\end{figure}

To evaluate the importance of considering velocity as an input, we group data from different velocities by time-scaling the interaction (under the quasi-static assumption, an action executed twice as fast will have the same outcome, except in half the time) and train models without the velocity as input. 
If the system is indeed quasi-static, combining the data from different velocities should increase the amount of training data and yield better (or not worse) performance. Otherwise, we should see the performance worsen.
\figref{fig:quasi-static} shows the evolution of the performance of the learned model as we add data from larger velocities, all time-scaled to 10mm/s. We can see that the performance peaks at around 50-80mm/s and degrades after that. We conclude that the quasi-static assumption does not hold after that, and hence there is value in adding velocity as an explicit input.

\bibliographystyle{IEEEtranN} 
{\footnotesize \bibliography{mb-icra17-pushing}} 

\end{document}